\pdfoutput=1

\documentclass[11pt]{article}

\usepackage{acl}

\usepackage{times}
\usepackage{latexsym}
\usepackage{multirow}
\usepackage{makecell}
\usepackage{booktabs}
\usepackage{graphicx}
\usepackage{listings}
\usepackage{xcolor}
\usepackage{hyperref}
\usepackage{float}
\definecolor{codegreen}{rgb}{0,0.6,0}
\definecolor{codegray}{rgb}{0.5,0.5,0.5}
\definecolor{codepurple}{rgb}{0.58,0,0.82}
\definecolor{backcolour}{rgb}{0.9843,0.9882,0.9609}

\lstdefinestyle{mystyle}{
    backgroundcolor=\color{backcolour},   
    commentstyle=\color{codegreen},
    keywordstyle=\color{magenta},
    numberstyle=\tiny\color{codegray},
    stringstyle=\color{codepurple},
    basicstyle=\ttfamily\footnotesize,
    breakatwhitespace=false,         
    breaklines=true,                 
    captionpos=b,                    
    keepspaces=true,                 
    numbers=left,                    
    numbersep=5pt,                  
    showspaces=false,                
    showstringspaces=false,
    showtabs=false,                  
    tabsize=2
}
\usepackage[T1]{fontenc}

\usepackage[utf8]{inputenc}

\usepackage{microtype}

\title{M-SENA: An Integrated Platform for Multimodal Sentiment Analysis}

\author{
  Huisheng Mao\textsuperscript{\rm 1, 2}\thanks{\quad  These authors contributed equally to this work.}, Ziqi Yuan\textsuperscript{\rm 1, 2}\footnotemark[1], Hua Xu\textsuperscript{\rm 1, 2}\thanks{\quad  Hua Xu is the corresponding author.}, Wenmeng Yu\textsuperscript{\rm 1, 2}, Yihe Liu\textsuperscript{\rm 1, 3}, Kai Gao\textsuperscript{\rm 3}\\
  \textsuperscript{\rm 1}State Key Laboratory of Intelligent Technology and Systems, \\ 
Department of Computer Science and Technology, Tsinghua University\\
  \textsuperscript{\rm 2}Beijing National Research Center for Information Science and Technology(BNRist)\\
  \textsuperscript{\rm 3}School of Information Science and Engineering, Hebei University of Science and Technology\\
    \texttt{\{mhs20,yzq21\}@mails.tsinghua.edu.cn}\\
    \texttt{xuhua@tsinghua.edu.cn}\\
}

\begin{document}
\maketitle
\begin{abstract}

M-SENA is an open-sourced platform for Multimodal Sentiment Analysis. It aims to facilitate advanced research by providing flexible toolkits, reliable benchmarks, and intuitive demonstrations. The platform features a fully modular video sentiment analysis framework consisting of data management, feature extraction, model training, and result analysis modules. In this paper, we first illustrate the overall architecture of the M-SENA platform and then introduce features of the core modules. Reliable baseline results of different modality features and MSA benchmarks are also reported. Moreover, we use model evaluation and analysis tools provided by M-SENA to present intermediate representation visualization, on-the-fly instance test, and generalization ability test results. The source code of the platform is publicly available at \url{https://github.com/thuiar/M-SENA}. 
\end{abstract}

\section{Introduction}
\label{sec: intro}
Multimodal Sentiment Analysis (MSA) aims to judge the speaker's sentiment from video segments \cite{MSA, MSASurvey, MSASurvey2}. It has attracted increasing attention due to the booming of user-generated online content. Although impressive improvements have been witnessed in recent MSA researches \cite{MulT, BERT_MAG, self-mm}, building an end-to-end video sentiment analysis system for real-world scenarios is still full of challenges.

The first challenge lies in effective acoustic and visual feature extraction. Most previous approaches \cite{TFN, MISA, han2021bi} are developed on the provided modality sequences from CMU-MultimodalSDK\footnote{\href{http://immortal.multicomp.cs.cmu.edu/raw_datasets/processed_data/}{Features provided by CMU}}. However, reproducing exact identical acoustic and visual feature extraction is almost impossible due to the the vague description of feature selection and backbone selection (both COVAREP\footnote{\href{https://github.com/covarep/covarep}{https://github.com/covarep/covarep}} and Facet\footnote{\href{https://imotions.com}{https://imotions.com}} can not be directly used in Python). Moreover, recent literature \cite{MulT, gkoumas2021makes, MMIM} observe that the text modality stands in the predominant position while acoustic and visual modalities have few contributions to the final sentiment classification. Such results further arouse the attention on effective feature extraction of acoustic and visual modalities.

With the awareness of the importance of acoustic and visual feature extraction, researchers attempt to develop models based on customized modality sequences instead of provided features \cite{dai2021multimodal, MISA}. However, performance comparison with different modality features is unfair. Therefore, the demand for reliable comparison of modality features and fusion methods is increasingly urgent. 

Another factor that limits the application of existing MSA models in real scenarios is the lack of comprehensive model evaluation and analysis approaches. Models obtained outstanding performance on the given test set might degrade in real-world scenarios due to the distribution discrepancy or random modality perturbations \cite{liang2019learning, zhao2021missing, yuan2021transformer}. Besides, effective model analysis is also crucial for researchers to explain the improvements and perform model refinement.

\begin{figure*}
  \centering
  \includegraphics[width=0.98\linewidth]{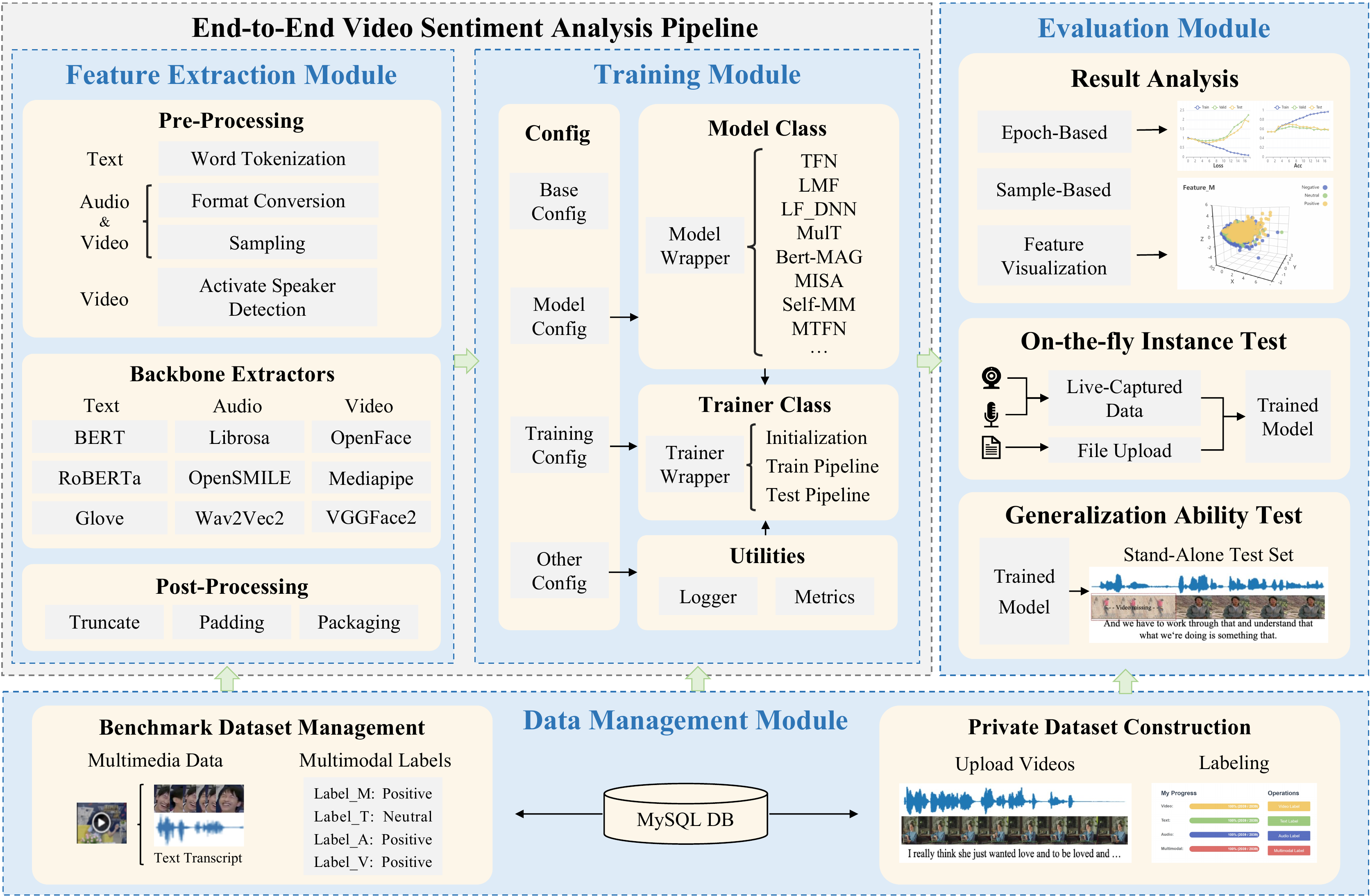}
  \caption{The overall framework of the M-SENA platform contains four main modules: data management module, feature extraction module, model training module and model evaluation module.}
  \label{fig: model}
\end{figure*}

The \textbf{M}ultimodal \textbf{SEN}timent \textbf{A}nalysis platform (M-SENA) is developed to address the above challenges. For acoustic and visual features, the platform integrates Librosa \cite{LIBROSA}, OpenSmile \cite{OPENSMILE}, OpenFace \cite{OPENFACE}, MediaPipe \cite{MEDIAPIPE} and provides a highly customized feature extraction API in Python. With the modular MSA pipeline, fair comparison between different features and MSA fusion models can be achieved. The results can be regarded as reliable baselines for future MSA research. Furthermore, the platform provides comprehensive model evaluation and analysis tools to reflect the model performance in real-world scenarios, including intermediate result visualization, on-the-fly instance demonstration, and generalization ability test. The contributions of this work are briefly summarized as follows:
\begin{enumerate}
    \item By providing a highly customized feature extraction toolkit, the platform familiarizes researchers with the composition of modality features. Also, the platform bridges the gap between designing MSA models with provided, fixed modality features and building a real-world video sentiment analysis system.
    \item The unified MSA pipeline guarantees fair comparison between different combinations of modality features and fusion models.
    \item To help researchers evaluate and analyze MSA models, the platform provides tools such as intermediate result visualization, on-the-fly instance demonstration, and generalization ability test.
\end{enumerate}

\section{Platform Architecture}
\label{sec: architecture}

M-SENA platform features convenient data access, customized feature extraction, unified model training pipeline, and comprehensive model evaluation. It provides a graphical web interface as well as Python packages for researchers with all features above. The platform currently supports three popular MSA datasets across two languages, seven feature extraction backbones, and fourteen benchmark MSA models. Figure \ref{fig: model} illustrates the overall architecture of the M-SENA platform. In the remaining parts of this section, features of each module in Figure \ref{fig: model} will be described in detail.

\subsection{Data Management Module}

The data management module is designed to ease the access of multimedia data on servers. Besides providing existing benchmark datasets, the module also enables researchers to build and manage their own datasets.

\noindent\textbf{Benchmark Datasets.} M-SENA currently supports three benchmark MSA datasets, including CMU-MOSI \cite{MOSI}, CMU-MOSEI \cite{MOSEI} in English, and CH-SIMS \cite{CH-SIMS} in Chinese. Details of integrated datasets are shown in Appendix \ref{sec: datasets}. Users can filter and view raw videos conveniently without downloading them to the local environment. 

\noindent\textbf{Building Private Datasets.} The M-SENA platform also provides a graphical interface for researchers to construct their own datasets using uploaded videos. Following the literature \cite{CH-SIMS}, M-SENA supports unimodal sentiment labelling along with multimodal sentiment labelling. The constructed datasets can be directly used for model training and evaluation on the platform.

\subsection{Feature Extraction Module}

Emotion-bearing modality feature extraction is still an open challenge for MSA tasks. To facilitate effective modality feature extraction for MSA, M-SENA integrates seven most commonly used feature extraction tools and provides a unified Python API as well as a graphical interface. Part of the supported features for each modality are listed in Table \ref{tab: features} and described below:

\begin{table}
\small
\centering
\begin{tabular}{lc}
\toprule[1pt]
\multicolumn{2}{c}{Acoustic Feature Sets}  \\
\midrule[1pt]
ComParE\_2016 \cite{compare_2016} & Static (HSFs) \\
eGeMAPS \cite{egemaps} & Static (LLDs) \\
wav2vec2.0 \cite{wav2vec2} & Learnable \\
\midrule[0.5pt]
\midrule[0.5pt]
\multicolumn{2}{c}{Visual Feature Sets}  \\
\midrule[1pt]
Facial Landmarks \cite{landmarks} & Static\\ 
Eyes Gaze \cite{wood2015rendering} & Static\\ 
Action Unit \cite{action_unit} & Static\\
\midrule[0.5pt]
\midrule[0.5pt]
\multicolumn{2}{c}{Textual Feature Sets} \\
\midrule[1pt]
GloVe6B \cite{GLOVE} & Static \\
BERT \cite{BERT} & Learnable\\
RoBerta \cite{ROBERTA} & Learnable \\
\bottomrule[1pt]
\end{tabular}
\caption{Some of the supported features in M-SENA.}
\label{tab: features}
\end{table}

\noindent\textbf{Acoustic Modality.} Various acoustic features have been proven effective for emotion recognition \cite{el2011survey, akccay2020speech}. Hand-crafted acoustic features can be divided into two classes, low level descriptors (LLDs), and high level statistics functions (HSFs). LLDs features, including prosodies, spectral domain features and others, are calculated on a frame-basis, while HSFs features are calculated on an entire utterance level. In addition to the hand-crafted features, M-SENA also provides pretrained acoustic model wav2vec2.0 \cite{wav2vec2} as a learnable feature extractor. Researchers can also design and build their own customized acoustic features using the provided Librosa extractor. 

\noindent\textbf{Visual Modality.} In existing MSA research, facial Landmarks, eyes gaze, and facial action units are commonly used visual features. The M-SENA platform enables researchers to extract visual feature combinations flexibly using OpenFace and MediaPipe extractors.

\noindent\textbf{Text Modality.} Compared with acoustic and visual features, semantic text embeddings are much more mature with the rapid development of pretrained language models \cite{pretrained_survey}. Following previous works \cite{TFN, BERT_MAG, roberta_1}, M-SENA supports GloVe6B \cite{GLOVE}, pretrained BERT \cite{BERT}, and pretrained RoBerta \cite{ROBERTA} as textual feature extractors.

All feature extractors above are available through both Python API and Graphical User Interface(GUI). Listing \ref{lst:mmsa-fet} shows a simple example of default acoustic feature extraction using Python API. The process is similar for other modalities. Advanced usage and detailed documentation is available at Github Wiki\footnote{\href{https://github.com/thuiar/MMSA-FET/wiki}{https://github.com/thuiar/MMSA-FET/wiki}}. 

\begin{minipage}{0.94\linewidth}

\begin{lstlisting}[language=Python, caption={An example of acoustic feature extraction on the MOSI dataset using MMSA.}, captionpos=b, label={lst:mmsa-fet}]
from MSA_FET import FeatureExtractionTool

# Extract Audio Feature for MOSI.
fet = FeatureExtractionTool("librosa")

feature = fet.run_dataset(
    dataset_dir='~/MOSI', 
    out_file='output/feature.pkl'
)

\end{lstlisting}
\end{minipage}

\begin{table}[t]
  \centering
  \scriptsize
  \begin{tabular}{cccc}
  \toprule[1pt]
  \multirow{2}{*}{Types} & \multicolumn{3}{c}{Scenarios}  \\ 
  \cline{2-4}
                    & Films(TV)   & Variety Show   & Life(Vlog)  \\ 
  \midrule[1pt]
  Easy              & 10 (en:4 ch:6)       & 8 (en:4 ch:4)     & 8 (en:4 ch:4)    \\
  Common            & 9 (en:4 ch:5)       & 11 (en:6 ch:5)     & 8 (en:4 ch:4)    \\
  Difficult         & 9 (en:4 ch:5)       & 9 (en:5 ch:4)     & 8 (en:4 ch:4)    \\
  Noise             & 9 (en:4 ch:5)       & 8 (en:4 ch:4)     & 7 (en:2 ch:5)    \\
  Missing           & 9 (en:4 ch:5)       & 9 (en:5 ch:4)     & 7 (en:3 ch:4)   \\
  \bottomrule[1pt]
  \end{tabular}
  \caption{Statistics of the generalization ability test dataset, where "en" represents "English", "ch" represents "Chinese".}
  \label{tab:generalization}
\end{table}

\begin{table*}
\small
\centering
\begin{tabular}{c|cc|cc|cc|cc}
\toprule[1pt]
\multirow{2}{*}{Feature Combinations} & \multicolumn{2}{c|}{TFN} & \multicolumn{2}{c|}{GMFN} & \multicolumn{2}{c|}{MISA} & \multicolumn{2}{c}{Bert-MAG}  \\ 
  & Acc-2 (\%) & F1 (\%) & Acc-2 (\%) & F1 (\%) & Acc-2 (\%) & F1 (\%) & Acc-2 (\%) & F1 (\%)\\ 
\cmidrule[1pt]{1-9}
CMU-SDK$^{\dag}$ & 78.02 & 78.09 & 76.98 & 77.06 & 82.96 & 82.98 & 83.41 & 83.47\\
 
[T1]-[A1]-[V1] & 77.41 & 77.47 & 77.77 & 77.84 & 83.78 & 83.80 & 83.38 & 83.43 \\
\cmidrule[1pt]{1-9}
[T2]-[A1]-[V1] & 70.40 & 70.51 & 71.40 & 71.54 & 75.22 & 75.68 & - & - \\ 

[T3]-[A1]-[V1] & 80.85 & 80.79 & 80.21 & 80.15 & 79.57 & 79.67 & - & - \\
\cmidrule[1pt]{1-9}
[T1]-[A2]-[V1] & 76.80 & 76.82 & 78.02 & 78.03 & 83.72 & 83.72 & 82.96 & 83.04 \\

[T1]-[A3]-[V1] & 77.19 & 77.23 & 78.44 & 78.45 & 82.16 & 82.23 & 83.57 & 83.58 \\
\cmidrule[1pt]{1-9}
[T1]-[A1]-[V2] & 77.38 & 77.48 & 78.81 & 78.71 & 83.2 & 83.14 & 82.13 & 82.20 \\

[T1]-[A1]-[V3] & 76.74 & 76.81 & 78.23 & 78.24 & 84.06 & 84.08 & 83.69 & 83.75 \\
\bottomrule[1pt]
\end{tabular}
\caption{Results for feature selection. For text, [T1] refers to BERT, [T2] refers to GloVe6B, [T3] refers to RoBerta. For acoustic, [A1] refers to eGeMAPS, [A2] refers to customized feature including 20-dim MFCC, 12-dim CQT, and f0, [A3] refers to wav2vec2.0. For visual, [V1] refers to action units, [V2] refers to landmarks, [V3] refers to both landmarks and action units. CMU-SDK$^{\dag}$ refers to modified CMU-SDK features with BERT for text. }
\label{tab: result}
\end{table*}

\subsection{Model Training Module}

M-SENA provides a unified training module which currently integrates 14 MSA benchmarks, including tensor fusion methods, TFN \cite{TFN}, LMF \cite{LMF}, modality factorization methods,  MFM \cite{MFM}, MISA \cite{MISA}, SELF-MM \cite{self-mm}, word-level fusion methods, MulT \cite{MulT}, BERT-MAG \cite{BERT_MAG}, multi-view learning methods: MFN \cite{MFN}, GMFN \cite{MOSEI}, and other MSA methods. Detailed introduction of the integrated baseline methods is provided in Appendix \ref{sec: benchmarks}. We will continue following advanced MSA benchmarks and put our best effort into providing reliable benchmark results for future MSA research.

\subsection{Result Analysis Module}
The proposed M-SENA platform provides comprehensive model evaluation tools including intermediate result visualization, on-the-fly instance test, and generalization ability test. A brief introduction of each component is given below, while a detailed demonstration is shown in Section \ref{sec: evaluation}.

\noindent\textbf{Intermediate Result Visualization.} The discrimination of multimodal representations is one of the crucial metrics for the evaluation of different fusion methods. The M-SENA platform records the final multimodal fusion results and illustrates them after decomposition with Principal Component Analysis (PCA). Training loss, binary accuracy, F1 score curves are also provided in M-SENA for detailed analysis.

\noindent\textbf{Live Demo Module.} In the hope of bridging the gap between MSA research and real-world video sentiment analysis scenarios, M-SENA provides a live demo module, which performs on-the-fly instance tests. Researchers can validate the effectiveness and robustness of the selected MSA model by uploading or live-feeding videos to the platform. 
  
\noindent\textbf{Generalization Ability Test.} Compared to the provided test set of benchmark MSA datasets, real-world scenarios are often more complicated. Future MSA models need to be robust against modality noise as well as effective on the test set. Driven by the demand from real-world applications and observations, the M-SENA platform provides a generalization ability test dataset (consists of 68 Chinese and 61 English samples), simulating as many complicated and diverse real-world scenarios as possible. The statistics of the proposed dataset is shown in Table \ref{tab:generalization}. In general, the dataset contains three scenarios and five instance types. Specifically, the three scenarios refers to films, variety shows, and user-uploaded vlogs, while the five instance types refer to easy samples, common samples, difficult samples, samples with modality noise, samples with modality missing. In addition, the dataset is balanced in terms of gender 
and scenario to avoid irrelevant factors. Examples of the generalization ability test dataset are shown in Appendix \ref{sec: gat_data}.

\section{Experiments on M-SENA}
In this section, we report experiments conducted on the M-SENA platform. Comparison of different modality features are shown in Section \ref{sec: feature_comparison}, and comparison of different fusion models are shown in Section \ref{sec: model_comparison}. All reported results are the mean performances of five different seeds.

\begin{table*}[t]
  \centering
  \small
  \begin{tabular}{c||cccc|cccc|cccc}
  \toprule[1pt]
  \multirow{2}{*}{Model} & \multicolumn{4}{c|}{MOSI} & \multicolumn{4}{c|}{MOSEI} & \multicolumn{4}{c}{SIMS} \\ 
                   & Acc-2 &   F1  &  MAE  &  Corr & Acc-2 &   F1  &  MAE  & Corr  & Acc-2 & F1    &  MAE  & Corr  \\ 
  \midrule[1pt]
  LF\_DNN          & 79.39 & 79.45 & 0.945 & 0.675 & 82.78 & 82.38 & 0.558 & 0.731 & 76.68 & 76.48 & 0.446 & 0.567 \\
  EF\_LSTM         & 77.35 & 77.43 & 0.995 & 0.644 & 81.23 & 81.02 & 0.588 & 0.695 & 69.37 & 56.82 & 0.591 & 0.380 \\
  TFN              & 78.02 & 78.09 & 0.971 & 0.652 & 82.23 & 81.47 & 0.573 & 0.718 & 77.07 & 76.94 & 0.437 & 0.582 \\
  LMF              & 78.60 & 78.61 & 0.934 & 0.663 & 83.83 & 83.68 & 0.562 & 0.735 & 77.42 & 77.35 & 0.438 & 0.578 \\
  MFN              & 78.78 & 78.71 & 0.938 & 0.665 & 83.30 & 83.23 & 0.570 & 0.720 & 78.55 & 78.23 & 0.442 & 0.575 \\
  GMFN             & 76.98 & 77.06 & 0.986 & 0.642 & 83.48 & 83.23 & 0.575 & 0.713 & 78.77 & 78.21 & 0.445 & 0.578 \\
  MFM              & 78.63 & 78.63 & 0.958 & 0.649 & 83.49 & 83.29 & 0.581 & 0.721 & 75.06 & 75.58 & 0.477 & 0.525 \\
  MulT             & 80.21 & 80.22 & 0.912 & 0.695 & 84.63 & 84.52 & 0.559 & 0.733 & 78.56 & 79.66 & 0.453 & 0.564 \\
  MISA             & 82.96 & 82.98 & 0.761 & 0.772 & 84.79 & 84.73 & 0.548 & 0.759 & 76.54 & 76.59 & 0.447 & 0.563 \\
  BERT\_MAG        & 83.41 & 83.47 & 0.761 & 0.776 & 84.87 & 84.85 & 0.539 & 0.764 & 74.44 & 71.75 & 0.492 & 0.399 \\
  MLF\_DNN         &   -   &   -   &   -   &   -   &   -   &   -   &   -   &   -   & 80.44 & 80.28 & 0.396 & 0.665 \\
  MTFN             &   -   &   -   &   -   &   -   &   -   &   -   &   -   &   -   & 81.09 & 81.01 & 0.395 & 0.666 \\
  MLMF             &   -   &   -   &   -   &   -   &   -   &   -   &   -   &   -   & 79.34 & 79.07 & 0.409 & 0.639 \\
  Self\_MM         & 84.30 & 84.31 & 0.720 & 0.793 & 84.06 & 84.12 & 0.531 & 0.766 & 80.04 & 80.44 & 0.425 & 0.595 \\
  
  \bottomrule[1pt]
  \end{tabular}
  \caption{Experiment results for MSA benchmark comparison. All models utilize the Bert embedding and the provided acoustic and visual features in CMU-MultimodalSDK. Due to the requirement of unimodal labels, multi-task models, including MLF\_DNN, MTFN, and MLMF, are tested on SIMS only.}
  \label{res:model-dataset}
\end{table*}

\subsection{Feature Selection Comparison}
\label{sec: feature_comparison}

In the following experiments, we take BERT [T1], eGeMAPS (LLDs) [A1], and Action Unit [V1] as default modality features, and compare them with the other six feature sets. Specifically, we utilize GloVe6B [T2], RoBerta [T3] for text modality comparison; customized acoustic feature[A2](including 20 dimensional MFCC, 12 dimensional CQT, and 1 dimensional f0), wav2vec2.0 features [A3] for acoustic modality comparison; facial landmarks [V2], facial landmarks and action units [V3] for visual modality comparison. Besides, we also report the model performances using the modality features provided in CMU-MultimodalSDK. 

Table \ref{tab: result} shows the experiment results for feature selection. For Bert-MAG which is designed upon the Bert backbone, experiments are conducted only for Bert as text feature. It can be observed that, in most cases, using appropriate features instead of original features in CMU-MultimodalSDK helps to improve model performance. For textual modality, Roberta feature performs best for TFN and GMFN model, while Bert feature performs best for MISA model. For acoustic modality, wav2vec2.0 embeddings (without finetune) perform best for GMFN and Bert-MAG model. According to literature \cite{chen2021exploring, pepino2021emotion}, finetuning wav2vec2.0 can further improve model performance which might provide more effective acoustic features for future MSA research. For Visual modality, the combination of facial landmarks and action units achieves the overall best result, revealing the effectiveness of both landmarks and action units for sentiment classification.

\subsection{MSA Benchmark Comparison}
\label{sec: model_comparison}
Experiment results of benchmark MSA models are shown in Table \ref{res:model-dataset}. All models are improved using Bert as text embeddings while using original acoustic and visual features provided in CMU-MultimodalSDK. Besides recording reliable benchmark results, the M-SENA platform also provides researchers with a convenient approach to reproduce the benchmarks. Again, both GUI and Python API are available. We show an example of the proposed Python API in Listing \ref{lst:mmsa}. Detailed and Advanced usage is included in our documentation at Github\footnote{\href{https://github.com/thuiar/MMSA/wiki}{https://github.com/thuiar/MMSA/wiki}}. We will continuously catch up on new MSA approaches and update their performances.

\begin{minipage}{0.94\linewidth}

\begin{lstlisting}[language=Python, caption={An example to train model with M-SENA.}, captionpos=b, label={lst:mmsa}]
from MMSA import MMSA_run

# Load Default Training Config.
config = get_config_regression(
    model_name='tfn', 
    dataset_name='mosi'
)

# Using User Designed Hyper-parameter.
config['post_fusion_dim'] = 32

# Modality Feature Selection.
config['featurePath'] = 'feature.pkl'

# Start Model Training.
MMSA_run(
    model_name='tfn',
    dataset_name='mosi', 
    config=config,
    seeds=[1111]
)

\end{lstlisting}
\end{minipage}



\section{Model Analysis Demonstration}
\label{sec: evaluation}
This section demonstrates model analysis results using the M-SENA platform. Intermediate result analysis is presented in Section \ref{sec: intermediate_result_test}, on-the-fly instance analysis is shown in Section \ref{sec: on_the_fly}, and generalization ability analysis is illustrated in Section \ref{sec: gat}.

\begin{figure}[t]
  \centering
  \includegraphics[width=\linewidth]{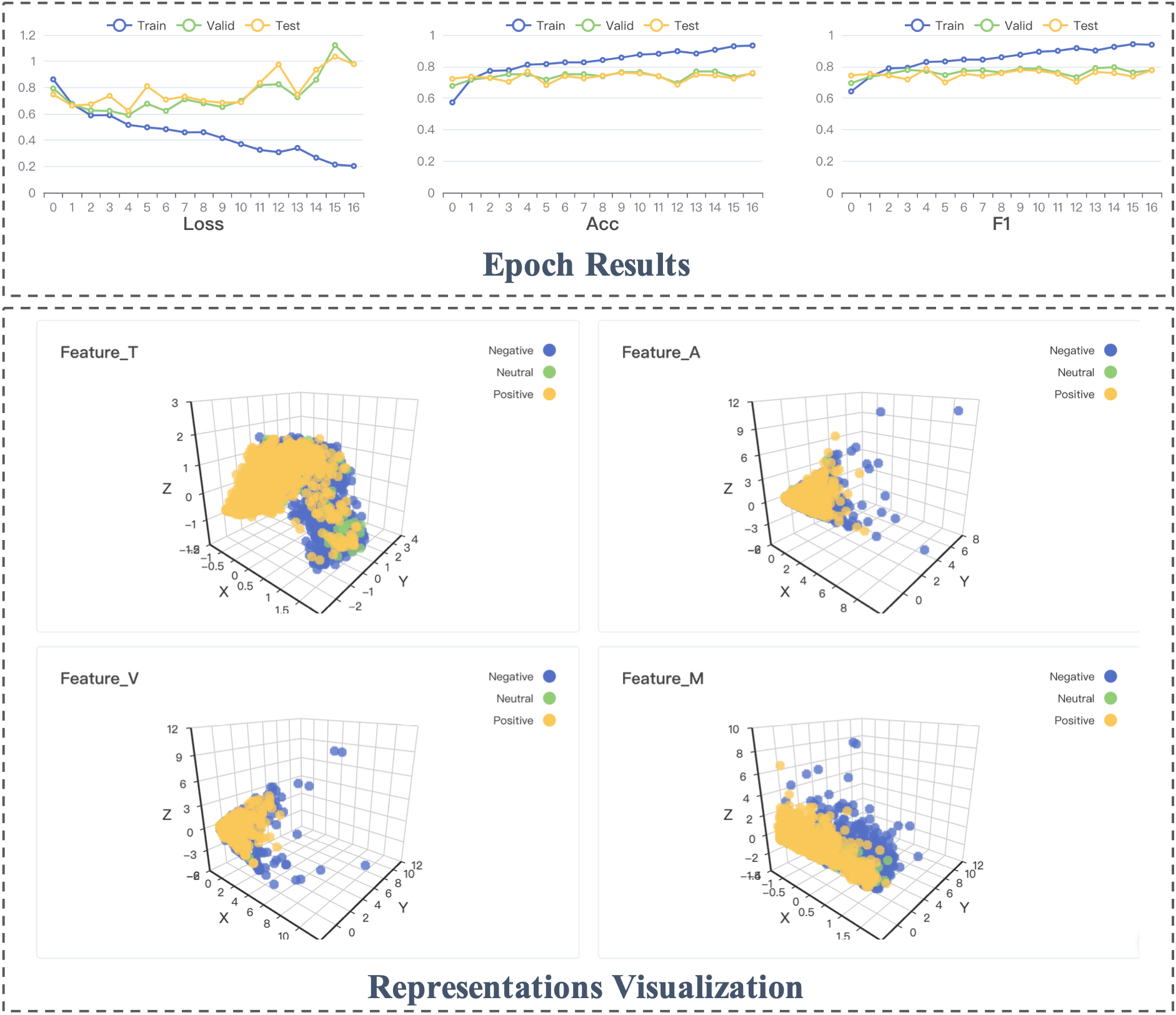}
  \caption{Intermediate Result Analysis for TFN model trained on MOSI dataset.}
  \label{fig: intermediate_result_test}
\end{figure}

\subsection{Intermediate Result Analysis}
\label{sec: intermediate_result_test}
The intermediate result analysis submodule is designed to monitor and visualize the training process. Figure \ref{fig: intermediate_result_test} shows an example of training TFN model on MOSI dataset. Epoch results of binary accuracy, f1-score and loss value are plotted. Moreover, the learned multimodal fusion representations are illustrated in an interactive 3D figure with the aim of helping users gain a better intuition about the multimodal feature representations and the fusion process. Unimodal representations of text, acoustic, and visual are also shown for models containing explicit unimodal representations.

\begin{figure}
  \centering
  \includegraphics[width=\linewidth]{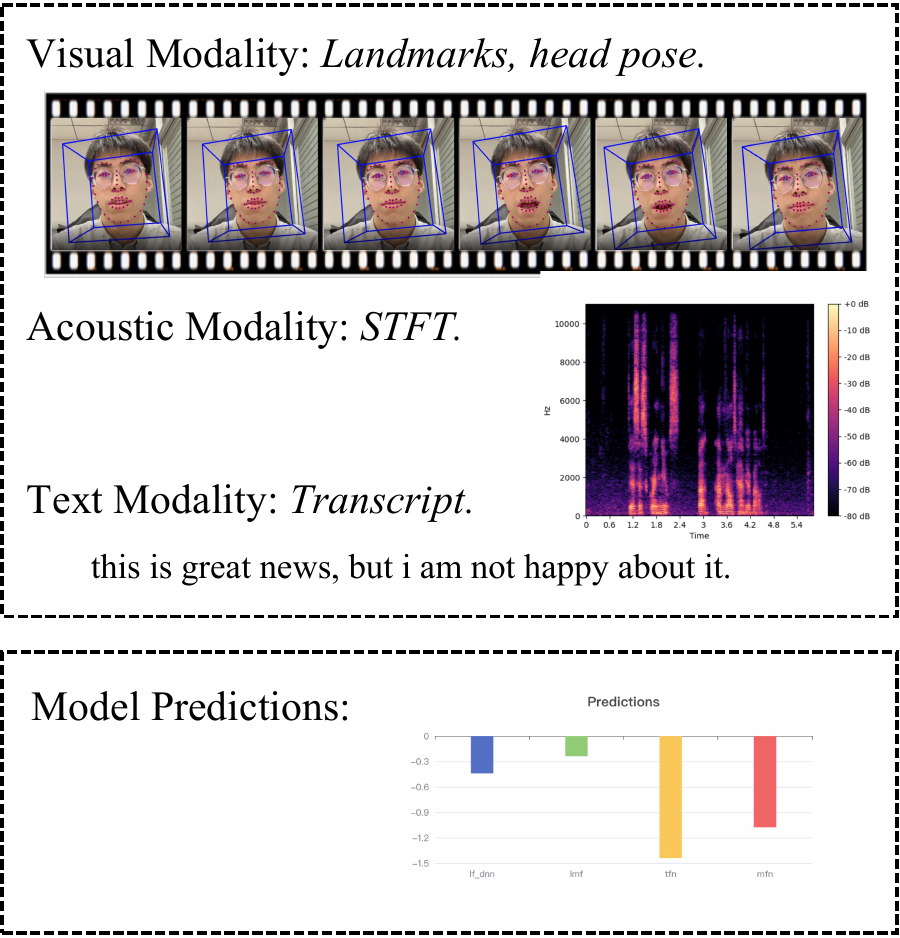}
  \caption{On-the-fly instance test example. The M-SENA platform also provides real-time modality feature visualization along with the model prediction results.}
  \label{fig: on-the-fly}
\end{figure}

\subsection{On-the-fly Instance Analysis}
\label{sec: on_the_fly}

M-SENA enables researchers to validate the proposed MSA approaches using uploaded or live-recorded instances. Figure \ref{fig: on-the-fly} presents an example of the live demonstration. Besides model prediction results, the platform also provides feature visualization, including short-time Fourier transform (STFT) for acoustic modality and facial landmarks, eye gaze, head poses for visual modality. We will continuously update the demonstration to make it a even more intuitive and playable MSA model evaluation tool.

\subsection{Generalization Ability Analysis}
\label{sec: gat}
We utilized the model trained on MOSI dataset with [T1]-[A1]-[V3] modality features in Section \ref{sec: feature_comparison} for generalization ability test. Experimental results are reported in Table \ref{tab: gat_result_en}. It can be concluded that all models present a performance gap between original test set and real-world scenarios, especially for the instances with noisy or missing modalities. Another observation is that the noisy instances are usually more challenging than modality missing for MSA models, revealing that noisy modality feature is worse than none at all. In the future, for the demand of real-world applications, MSA researchers may consider analyzing model robustness as well as performances on the test set, and design a more robust MSA model against random modality noise.

\begin{table}[t]
  \centering
  \scriptsize
  \begin{tabular}{cc|c|c|c}
  \toprule[1pt]
  \multirow{2}{*}{Types} & \multicolumn{1}{c|}{TFN} & \multicolumn{1}{c|}{GMFN} & \multicolumn{1}{c}{MISA}  & \multicolumn{1}{c}{Bert-MAG}\\ 
                   & Acc-2 / F1 & Acc-2 / F1 & Acc-2 / F1 & Acc-2 / F1\\
  \midrule[1pt]
  Easy              & 83.3 / 84.4 & 75.0 / 76.1 & 75.0 / 76.7 & 66.7 / 66.7   \\
  Common            & 71.4 / 74.5 & 85.7 / 82.3 & 71.4 / 75.8 & 78.6 / 78.6   \\
  Difficult         & 69.2 / 69.2 & 61.5 / 60.5 & 53.9 / 54.4 & 84.6 / 84.6   \\
  Noise             & 60.0 / 50.5 & 50.0 / 44.9 & 50.0 / 35.7 & 60.0 / 51.7   \\
  Missing           & 63.6 / 60.6 & 81.8 / 77.8 & 63.6 / 60.6 & 63.6 / 61.5   \\
  \midrule[1pt]
  Avg               & 70.0 / 68.4 & 71.7 / 69.3 & 63.3 / 62.4  & 71.7 / 69.7   \\
  \bottomrule[1pt]
  \end{tabular}
  \caption{Results for English generalization ability test. Binary accuracy and F1 scores are reported to show the effectiveness and robustness of the model.}
  \label{tab: gat_result_en}
\end{table}

\section{Related Works}

To the best of our knowledge, there are two widely used open-source repositories from CMU team\footnote{\href{https://github.com/A2Zadeh/CMU-MultimodalSDK}{https://github.com/A2Zadeh/CMU-MultimodalSDK}} and SUTD team\footnote{\href{https://github.com/declare-lab/multimodal-deep-learning}{https://github.com/declare-lab/multimodal-deep-learning}}. Both of them provide tools to load well-known MSA datasets and implement several benchmarks methods. So far, their works have attracted considerable attention and facilitated the birth of new MSA models such as MulT \cite{MulT} and MMIM \cite{MMIM}.

In this paper, we propose M-SENA, compared to previous works, the M-SENA platform is novel from the following aspects. For data management, previous work directly loads the extracted features, while the M-SENA platform focuses on intuitive raw video demonstration, and provides user with a convenient means for private dataset construction. For modality features, M-SENA platform first provides user-customized feature extraction toolkit and a transparent feature extraction process. Following the tutorial, Users can easily reproduce the feature extraction steps and develop their research on designed feature set. For model training, the M-SENA platform first utilizes a unified MSA framework and provide an easy-to-reproduce model training API integrating fourteen MSA benchmarks on three popular MSA dataset. For model evaluation, the M-SENA is the first MSA platform consisting of comprehensive evaluation means stressing model robustness for real-world scenarios, which aims to bridge the gap between MSA research and applications.

\section*{Conclusion}

In this work, we introduce M-SENA, an integrated platform that contains step-by-step recipes for data management, feature extraction, model training, and model analysis for MSA researchers. The platform evaluates MSA model in an end-to-end manner and reports reliable benchmark results for future research. Moreover, we further investigate comprehensive model evaluation and analysis methods and provide a series of user-friendly visualization and demonstration tools including intermediate representation visualization, on-the-fly instance test, and generalization ability test. In the future, we will continuously catch up on advanced MSA research progress and update new benchmarks on the M-SENA platform.

\section*{Acknowledgement}
This paper is funded by The National Natural Science Foundation of China (Grant No. 62173195) and Beijing Academy of Artificial Intelligence (BAAI). The authors thank the anonymous reviewers for their valuable suggestions.


\bibliographystyle{acl_natbib}

\appendix

\section{Integrated Datasets}
\label{sec: datasets}

\noindent \textbf{CMU-MOSI.} 
 The MOSI \cite{MOSI} dataset is a widely-used dataset that consists of a collection of 2,199 video segments from 93 YouTube movie review videos.  
 
\noindent \textbf{CMU-MOSEI.} 
The MOSEI \cite{MOSEI} dataset expands the MOSI dataset by enlarging the number of utterances and enriching the variety of samples, speakers, and topics. For both MOSI and MOSEI datasets, instances are annotated with a sentiment intensity score ranging from -3 to 3 (strongly negative to strongly positive).

\noindent \textbf{CH-SIMS.} 
The SIMS dataset \cite{CH-SIMS} is a Chinese unimodal and multimodal sentiment analysis dataset. It contains 2,281 refined video segments in the wild with both multimodal and independent unimodal annotations of a sentiment intensity score ranging from -1 to 1 (negative to positive, the score interval is 0.2).

\section{Integrated Benchmarks}
\label{sec: benchmarks}

\noindent \textbf{LF-DNN.} The Late Fusion Deep Neural Network \cite{LF-DNN} first extracts modality features separately and performs late fusion strategy for final predictions.

\noindent \textbf{EF-LSTM.} The Early Fusion Long-Short Term Memory \cite{LF-DNN} is based on input-level feature fusion and conducts Long-Short Term Memory (LSTM) to learn multimodal representations.

\noindent\textbf{TFN.} The Tensor Fusion Network (TFN) \cite{TFN} calculates a multi-dimensional tensor (based on outer product) to capture uni-, bi-, and tri-modal interactions.

\noindent\textbf{LMF.} The Low-rank Multimodal Fusion (LMF) \cite{LMF} is an improvement over TFN, where the low-rank multimodal tensors fusion technique is performed to improve efficiency.

\noindent\textbf{MFN.} The Memory Fusion Network (MFN) \cite{MFN} accounts for continuously modeling the view specific and cross-view interactions and summarizing them through time with a Multi-view Gated Memory.

\noindent \textbf{Graph-MFN.} The Graph Memory Fusion Network \cite{MOSEI} is an improvement of MFN, which can change the fusion structure dynamically to obtain the interaction between the modalities and improve the interpretability.

\noindent\textbf{MulT.} The Multimodal Transformer (MulT) \cite{MulT} extends multimodal transformer architecture with directional pairwise cross-modal attention which translates one modality to another using directional pairwise cross-attention.

\noindent\textbf{BERT-MAG.} The Multimodal Adaptation Gate for Bert (MAG-BERT) \cite{BERT_MAG} is an improvement over RAVEN on aligned data with applying multimodal adaptation gate at different layers of the BERT backbone.

\noindent \textbf{MISA.} The Modality-Invariant and -Specific Representations \cite{MISA} is made up of a combination of losses including similarity loss, orthogonal loss, reconstruction loss and prediction loss to learn modality-invariant and modality-specific representation.

\noindent \textbf{MFM.} The Multimodal Factorization Model \cite{MFM} is a robust model, which can learn multimodal-discriminative and modality-specific generative factors, then reconstructs missing reconstruct missing modalities by adjusting for independent factors.

\noindent \textbf{MLF\_DNN.} The Multi-Task Late Fusion Deep Neural Network \cite{CH-SIMS} first extracts modality features separately and performs late fusion strategy for final predictions through unimodal labels training.

\noindent \textbf{MTFN.}  The Multi-Task Tensor Fusion Network \cite{CH-SIMS}  calculates a multi-dimensional tensor (based on outer product) to capture uni-, bi-, and tri-modal interactions through unimodal labels training.
 
\noindent \textbf{MLMF.}  The Multi-Task Low-rank Multimodal Fusion \cite{CH-SIMS} is an improvement over MTFN, where low-rank multimodal tensors fusion technique is performed to improve efficiency through unimodal labels training.

\noindent \textbf{Self\_MM.}  The Self-Supervised Multi-Task Multimodal \cite{self-mm} design a label generation module based on the self-supervised learning strategy to acquire independent unimodal supervisions, which can balance the learning progress among different sub-tasks.

\section{Generalization Ability Test Datasets}
\label{sec: gat_data}
\begin{figure}[H]
  \centering
  \includegraphics[width=\linewidth]{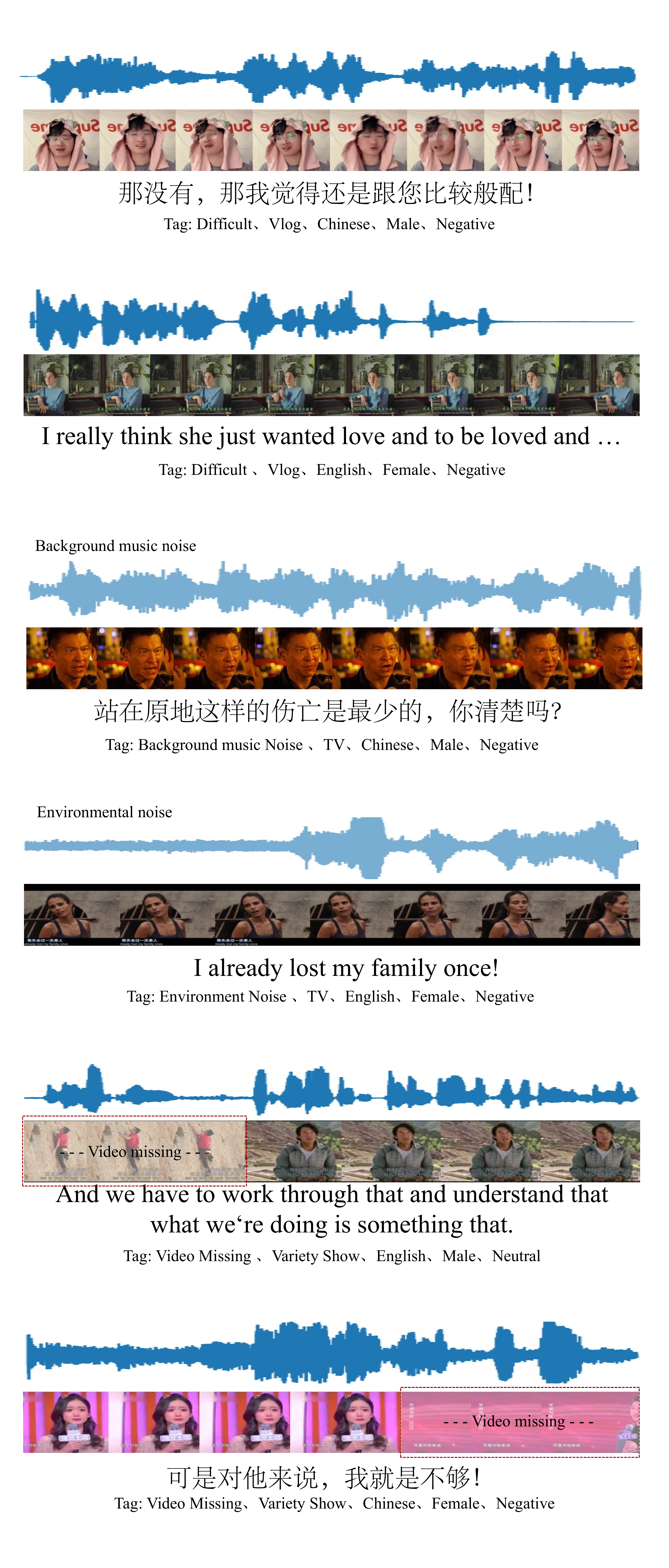}
  \caption{Examples of the constructed generalization ability test dataset.}
  \label{fig: gat_datasets_samples}
\end{figure}

The examples of the proposed generalization ability test dataset are shown in Figure \ref{fig: gat_datasets_samples}.

\newpage

\end{document}